\pdfoutput=1

\documentclass[11pt]{article}

\usepackage{ACL2023}
\usepackage{times}
\usepackage{latexsym}
\usepackage{xspace, subfigure}
\usepackage{amsmath}
\usepackage{amsfonts}
\usepackage{amssymb}

\usepackage{hyperref}
\usepackage{adjustbox}
\usepackage{array}
\usepackage{balance}
\usepackage{multirow}
\usepackage{babel, caption}
\newcommand{\our}{\text{ZS-SC}\xspace}
\newcommand\numberthis{\addtocounter{equation}{1}\tag{\theequation}}

\usepackage[T1]{fontenc}

\usepackage[utf8]{inputenc}

\usepackage{microtype}

\usepackage{inconsolata}

\makeatletter
\newcommand{\printfnsymbol}[1]{%
  \textsuperscript{\@fnsymbol{#1}}%
}
\makeatother

\begin{document}
\title{Zero-shot Approach to Overcome Perturbation Sensitivity of Prompts}
\author{Mohna Chakraborty\thanks{equal contribution}, Adithya Kulkarni\printfnsymbol{1}, \and Qi Li \\ Department of Computer Science, Iowa State University \\ \{mohnac, aditkulk, qli\}@iastate.edu}
\maketitle

\begin{abstract}
Recent studies have demonstrated that natural-language prompts can help to leverage the knowledge learned by pre-trained language models for the binary sentence-level sentiment classification task. Specifically, these methods utilize few-shot learning settings to fine-tune the sentiment classification model using manual or automatically generated prompts. However, the performance of these methods is sensitive to the perturbations of the utilized prompts. Furthermore, these methods depend on a few labeled instances for automatic prompt generation and prompt ranking. This study aims to find high-quality prompts for the given task in a zero-shot setting. Given a base prompt, our proposed approach automatically generates multiple prompts similar to the base prompt employing positional, reasoning, and paraphrasing techniques and then ranks the prompts using a novel metric. We empirically demonstrate that the top-ranked prompts are high-quality and significantly outperform the base prompt and the prompts generated using few-shot learning for the binary sentence-level sentiment classification task.
\end{abstract}
\section{Introduction}

The recent advance of large language models such as ChatGPT  \cite{chatgpt}, GPT-3 \cite{brown2020language}, and T5 \cite{raffel2020exploring} has shown an astounding ability to understand natural languages. These pre-trained models can conduct various Natural Language Processing (NLP) tasks under the zero/few-shot settings using natural language instructions (i.e., prompts) when no or a few training samples exist. The prompts play crucial roles in these scenarios.

The prompts can be generated manually or automatically \cite{schick2021exploiting, gao2021making, gu2022ppt, wang2022towards}. The manual prompts are handcrafted based on the user's intuition of the task \cite{schick2021exploiting, gao2021making}. Humans can easily write prompts, but the manual prompts are likely to be suboptimal since the language models may understand the instruction differently from humans. 
Prior studies have also shown that the performance of the language models is sensitive to the choice of prompts. For example, \cite{gao2021making, jiang2020can} have shown that the performance is sensitive to the choice of certain words in the prompts and the position of the prompts. Due to the sensitivity and the potential misunderstanding of the instruction, manual prompts tend to suffer from poor performance under zero-shot settings. The language models tend to understand human intentions better when used with a small amount of training data. Therefore, the model can improve significantly under few-shot settings.

To address the problems of manual prompts, some studies \cite{jiang2020can, gao2021making} further propose to generate prompts automatically following few-shot settings. These models utilize generative language models, such as the T5 model, to write automatic prompts using small training data from the task. Some studies \cite{shin2020autoprompt} also use the small training set to fine-tune the language models or to evaluate the prompts. However, there are several drawbacks to automatically generated prompts in real applications. First, prompts cannot be generated in zero-shot settings, and the generated prompts may not follow the human intuition of the tasks. Second, deploying the generative language models also poses challenges. It can be costly to deploy on local hardware due to the size of the pre-trained generative language models. Using the generative language models via API \cite{chatgpt} also faces limitations, such as privacy concerns when uploading confidential customer or organizational data. 

In this work\footnote{The code can be found at \url{https://github.com/Mohna0310/ZSSC}}, we aim to study how to improve manual prompts for classification tasks under zero-shot settings using moderately sized masked language models. Specifically, we use the binary sentence-level sentiment classification tasks as the testbed. Instead of deploying large generative language models, we study the usability of moderately sized masked language models, such as BERT \cite{devlin2019bert}, which can be deployed and tuned in-house easily for real-world applications. The prompt follows the cloze-style format, where the position of the label is masked (e.g., ``Battery life was great. The sentence was [MASK]'', where a positive polarity is the goal of prediction). The prompts are used to predict probability scores for the polarity labels from the pre-trained masked language model.

To overcome the sensitivity of the language model to a manual prompt, we propose augmentation strategies to automatically generate more candidate prompts similar to the manual prompt (i.e., the base prompt), which is not required to be complex or optimized. Three augmentation techniques are designed: positioning, subordination, and paraphrasing. Different from \citet{gao2021making}, where generative language models are used to generate candidate prompts, we use the same masked language models to paraphrase the base prompt. To find high-quality prompts under the zero-shot setting, we propose a novel ranking metric designed based on the intuition that high-quality prompts should be more sensitive to changing certain keywords. If a prompt is not sensitive to the change of certain keywords, it is not high-quality, and vice versa.  

We conduct extensive experiments on various benchmark datasets from different domains of binary sentence-level sentiment classification and show the efficacy of the proposed \our model compared with different prompts, including manually and automatically generated prompts, in the zero-shot setting. The experimental results demonstrate the effectiveness of the proposed method in real applications.

In summary, the main contributions of this paper are as follows:
\begin{itemize}
    \item We propose a prompt augmentation method using moderately sized masked language models to improve manual prompts for classification tasks under zero-shot settings. 
    \item To rank the automatically generated prompts under the zero-shot setting, we propose a novel ranking metric based on the intuition that high-quality prompts should be sensitive to the change of certain keywords in the given sentence.
    \item Extensive experiments and ablation studies performed on benchmark datasets for sentence-level sentiment classification tasks validate the effectiveness of the proposed method. 
\end{itemize} 
 
\section{Related Work}

Prompt-based learning is a recent paradigm used in the zero/few-shot setting. In the zero-shot setting, the model is given a natural language instruction (prompt) describing the task without any training data~\cite{brown2020language}, whereas in the few-shot setting, a few samples of training data are used along with the prompt. In prompt-based learning, the downstream tasks are formalized as masked language modeling problems using natural language prompts. Then, a verbalizer is used to map the masked language model prediction to the labels of the downstream task. This work uses prompt-based learning for the binary sentence-level sentiment classification task. This section discusses the related work that explored prompt-based learning from generic and task-specific perspectives.

\noindent \textbf{Prompt-based Learning}: With the introduction of GPT-3 \cite{brown2020language}, recent years have witnessed a series of studies based on prompt-based learning. \citet{schick2021exploiting} utilized manual-designed hard prompts, composed of discrete words, to fine-tune the pre-trained language model. Finding the best-performing manual prompt is challenging, and to alleviate the problem, \citet{jiang2020can, gao2021making, shin2020autoprompt} designed methods for automatic prompt generation. Specifically, \citet{shin2020autoprompt} performed the downstream tasks using gradient-guided search utilizing a large number of annotations for an automatic prompt generation. \citet{gao2021making} proposed LM-BFF that auto-generates prompts using the T5 model but relies on few annotations for an automatic prompt generation. However, the auto-generated prompts are hard prompts making them sub-optimal.  

To overcome the limitations of hard prompts, \citet{zhong2021factual, li2021prefix, wang2021transprompt} proposed methods to learn soft prompts under the few-shot settings. Soft (or continuous) prompts are composed of several continuous learnable embeddings, unlike hard prompts. Motivated by the prior studies, \citet{zhao2021discrete} utilized both the hard and soft prompts for training the pre-trained language model. \citet{gu2022ppt} proposed pre-training hard prompts by adding soft prompts into the pre-training stage to obtain a better initialization. 

Another line of study \cite{khashabi2022reframing, wang2022towards, zhong2021adapting} designed manual task-specific prompts by fine-tuning pre-trained language models on multiple tasks. The fine-tuned language model is then used on unseen tasks under the zero/few-shot setting.

\noindent \textbf{Prompt-based Learning for Sentence-level Sentiment Classification}:
Over the past years, a large body of studies \cite {shin2020autoprompt, gao2021making, gu2022ppt, wang2022towards} have demonstrated excellent performance in few-shot settings on sentence-level sentiment classification tasks. Specifically, \citet{shin2020autoprompt} used gradient-guided search to generate automatic prompts, whereas \citet{gao2021making} used a more general-purpose search method to generate automatic prompts. Following the limitation of automatic prompts, \citet{gu2022ppt} suggested hybrid training combining hard and soft prompts in the initial stage, obtaining a better initialization. \citet{wang2022towards} proposed a Unified Prompt Tuning framework and designed prompts by fine-tuning a pre-trained language model over a series of non-target NLP tasks and using the trained model to fit unseen tasks. For instance, when the target task is sentiment classification, the training data is from other domains like NLI and paraphrasing.

These studies consider access to labeled instances and perform the sentence-level sentiment classification task using a large-scale pre-trained generative language model. In our study, we do not use any training data, and the base prompt can be considered as a natural language description for the task. Therefore, this study follows the zero-shot setting. Using a moderately sized masked language model further makes the proposed method more appealing in practice.

\section{Methodology}
This section first discusses the problem formulation and the overview in Section \ref{problem formulation} and Section \ref{overview}. Our proposed method handles the language model's sensitivity to a manual prompt by utilizing prompt augmentation techniques to generate multiple candidate prompts. The detailed description of the prompt augmentation is discussed in Section \ref{prompt_augmentation}. To rank the automatically generated prompts in the zero-shot setting, we propose a novel ranking metric, discussed in Section \ref{ranking_metric}. Finally, the top-ranked prompts are used for prediction, discussed in Section \ref{inference}.

\subsection{Problem Formulation}
\label{problem formulation}
\begin{figure}[t]
    \centering
    \includegraphics[width=0.45\textwidth]{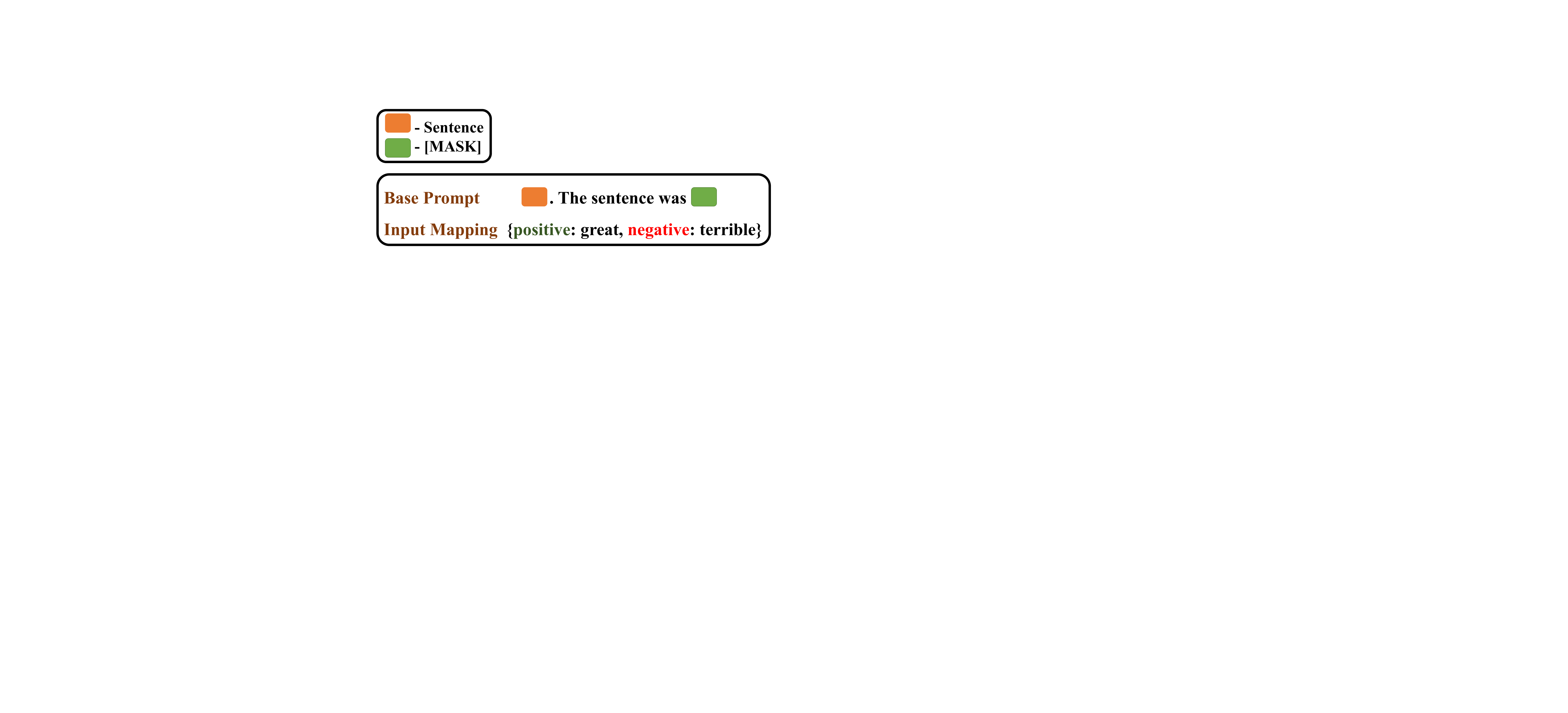}
    \caption{Model Input}
    \label{fig:input}
\end{figure}
Given an unlabeled corpus $\mathcal{D}$ with $N$ sentences, an input mapping $\mathcal{M}:\mathcal{Y}\rightarrow\mathcal{V}$ for the labels $y \in \mathcal{Y} = \{-1, 1\}$, in the vocabulary $\mathcal{V}$ of $\mathcal{L}$ and a base prompt $B_p$, the task is to find quality prompts similar to the base prompt in a zero-shot setting for the binary sentence-level sentiment classification task. 
Figure \ref{fig:input} shows one example input to the model. In this example, $y \in \mathcal{Y} = \{negative, positive\}$, $\mathcal{M}(positive)=great$, and $\mathcal{M}(negative)=terrible$.

\subsection{Overview}
\label{overview}

\begin{figure*}
    \centering
    \includegraphics[width=0.98\textwidth]{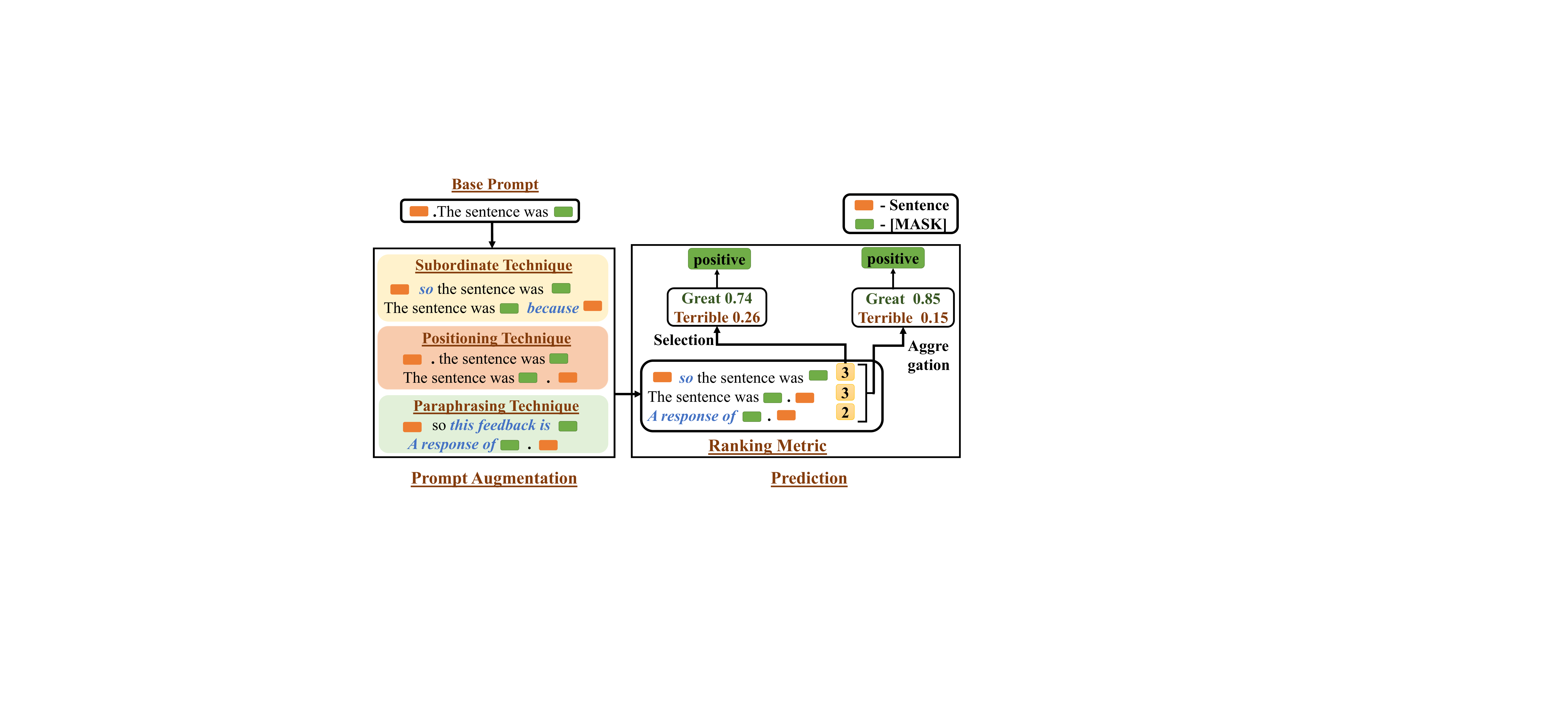}
    \caption{Overview of \our.}
    \label{fig:overview}
\end{figure*}

Given a base prompt $B_p$, the proposed \our first generates multiple prompts similar to the base prompt using augmentation techniques. Specifically, we introduce positioning, subordination, and paraphrasing techniques in the augmentation process, which are discussed in detail in Section \ref{prompt_augmentation}.

With more automatically generated candidate prompts, \our ranks the prompts using a novel ranking metric. This metric is designed based on the observation that quality prompts should flip the predicted label if $\mathcal{M}(y)$ present in the sentence is replaced with $\mathcal{M}(y')$, where $y \neq y'$, whereas the predicted label should stay the same if $\mathcal{M}(y)$ is replaced with its synonyms. Section \ref{ranking_metric} discusses the proposed ranking metric in detail.

Finally, the top-ranked prompt is selected, or top$-k$ highly ranked prompts are aggregated to conduct the zero-shot prediction for the unlabeled corpus $\mathcal{D}$ (Section \ref{inference}).

Figure \ref{fig:overview} illustrates the overview of the proposed approach, \our. 

\subsection{Prompt Augmentation}
\label{prompt_augmentation}

A single base prompt provided by a user may not provide optimal results for the given task. Prior studies \cite{gao2021making, jiang2020can} have shown that the performance of the prompts is sensitive to the choice of certain words and the position of the prompts, respectively. Furthermore, we observe that using subordinate conjunctions to join the prompt and sentence can improve the method's performance on some datasets since it introduces a dependency between the prompt and sentence, thereby leading the model to relate the predicted label with the context of the sentence. 
Based on the above observations, we propose to apply three augmentation techniques to generate prompts automatically, namely positioning, subordination, and paraphrasing techniques. 

The \textit{positioning} technique places the prompt either before or after the given sentence. The \textit{subordination} technique uses subordinate conjunctions like \textit{"because" and "so"} to join the prompt and the sentence. Specifically, the conjunction \textit{"because"} is used if the prompt is placed before the sentence, and the conjunction \textit{"so"} is used if the prompt is placed after the sentence.

\begin{figure}[t]
    \centering
    \includegraphics[width=0.45\textwidth]{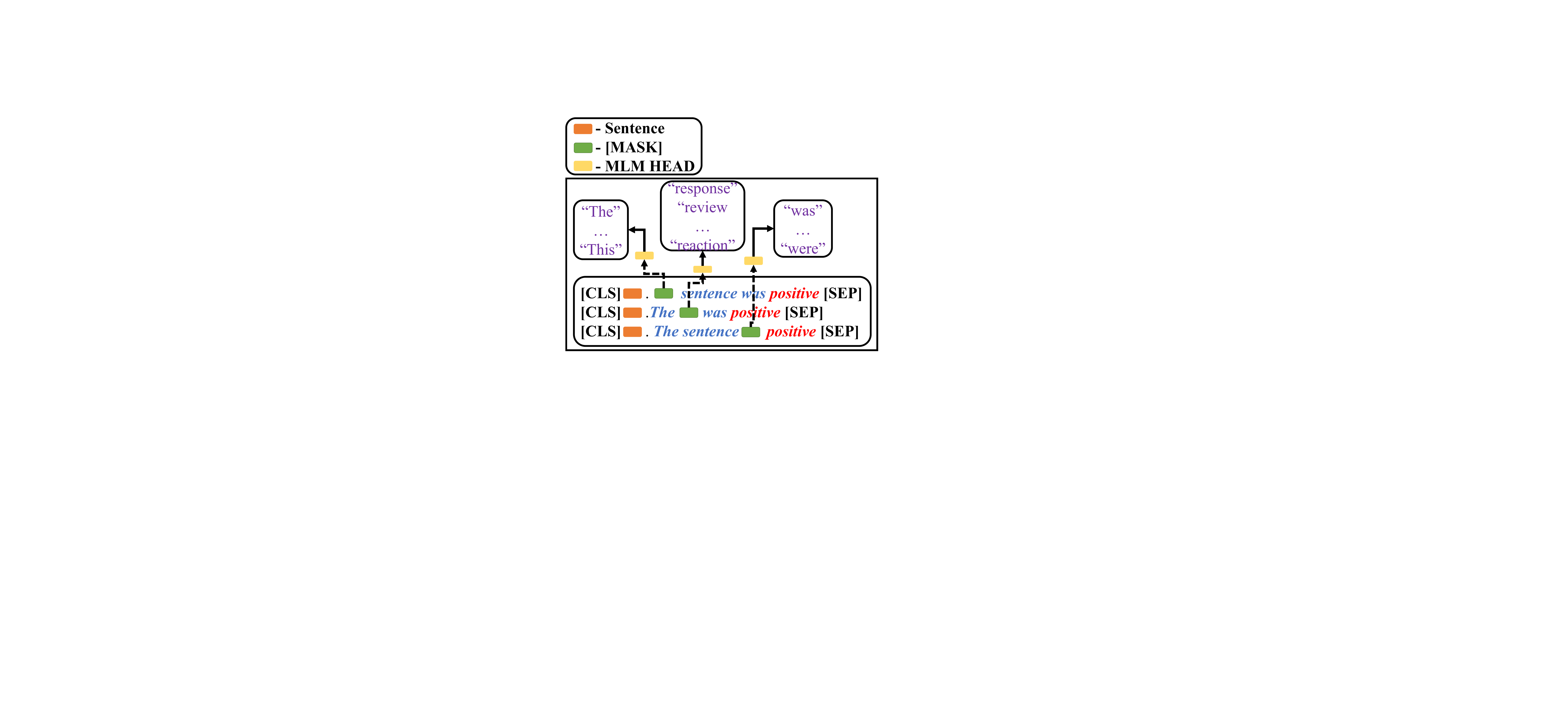}
    \caption{Paraphrasing Technique.}
    \label{fig:paraphrasing_technique}
\end{figure}
The \textit{paraphrasing} technique generates multiple prompts similar to the base prompt $B_p$ by swapping the tokens in the base prompt with similar tokens. These similar tokens should have the same part of speech tags as the tokens they are replacing and should not change the context of the prompt. Therefore, to obtain these similar tokens, we use a pre-trained MLM model $\mathcal{L}$. Pre-trained MLM models are trained to predict the missing tokens that fit the context of the given sentence and thus would be suitable for the purpose. Figure \ref{fig:paraphrasing_technique} illustrates the paraphrasing technique for the base prompt. The label ``positive" is used as a placeholder so that pre-trained MLM model can learn the context of the given sentence.

If a specific sentence is joined with the base prompt, the MLM model $\mathcal{L}$ can understand the context better, so the replacing tokens will make more sense. Therefore, instead of using prompts alone, we form sample instances by randomly selecting sentences from the unlabeled corpus $\mathcal{D}$.  
We then mask the replaceable tokens from the base prompt one at a time and use the MLM model $\mathcal{L}$ to predict the masked token. For each masked token, the MLM model $\mathcal{L}$ gives a score to all the tokens in its vocabulary. We choose the top-K ranked tokens as similar token candidates and remove those that do not have the same POS tag as the masked token.

These three techniques can be applied in different combinations and permutations to generate prompts automatically. The number of candidate paraphrasing tokens $K$ can be increased to generate more prompts. Figure \ref{fig:paraphrasing_technique} illustrates the process of obtaining paraphrasing tokens to the tokens of the base prompt.

\begin{figure}[t]
    \centering
    \includegraphics[width=0.48\textwidth]{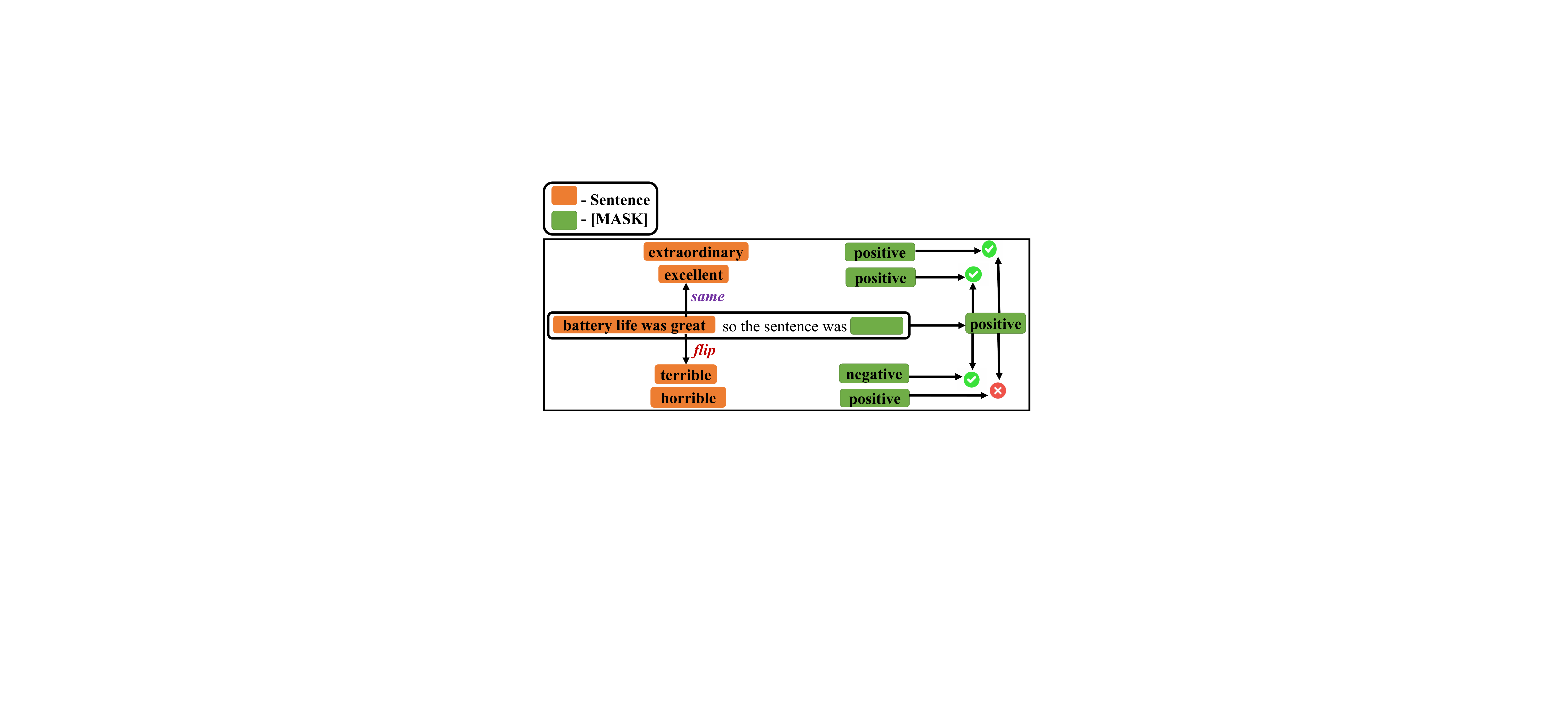}
    \caption{Ranking Metric.}
    \label{fig:ranking_metric}
\end{figure}

\subsection{Ranking Metric}
\label{ranking_metric}
Not all the automatically generated prompts in Section \ref{prompt_augmentation} obtain good performance for the task. Therefore, we aim to rank these prompts and choose quality prompts for the tasks. Previous works \cite{gao2021making, shin2020autoprompt} have used validation or manually annotated few-shot training data for evaluating the automatically generated prompts. However, under the zero-shot setting, we do not assume there exists any manually annotated data. Therefore, we have to rank the automatically generated prompts in the absence of manually annotated data which is not considered by the previous works.

Intuitively, if the mapping token of the opposite label replaces the mapping token in a given sentence, the predicted label by a quality prompt should flip. On the other hand, the predicted label should remain the same if the mapping token in the sentence is replaced by its synonyms. For example, suppose we replace the word \textit{"great"} in sentence \textit{"battery life was great"} with \textit{"terrible"}. In this case, the predicted label should flip, whereas if we replace \textit{"great"} with \textit{"excellent"}, the predicted label should remain the same. We use this intuition to measure the sensitivity of the prompt to the change of the mapping tokens in the given sentences. The measured sensitivity implies the quality of the prompt, namely prompts sensitive to the change of the mapping tokens in the given sentence can achieve good performance for the task. Figure \ref{fig:ranking_metric} illustrates the key idea of the proposed ranking metric.

We model the above intuition as a zero-one scoring function. To do so, we first obtain sentences from the unlabeled corpus $\mathcal{D}$ that contain the mapping tokens $\mathcal{M}(y) \in \mathcal{V}$ obtained from the provided input mapping $\mathcal{M}:\mathcal{Y}\rightarrow\mathcal{V}$. 
If the mapping tokens are not present in the corpus $\mathcal{D}$, the synonyms of the mapping tokens can be used. 

For a sentence $s_{in} \in S_{W}$, let the label predicted by the model for a given prompt $P$ be $l_1$. We then replace the mapping token $\mathcal{M}(y)$ in $s_{in}$ with $\mathcal{M}(y')$, where $y \neq y'$ to obtain a new sentence $s'_{in}$. Let the label predicted for $s'_{in}$ be $l_2$. The zero-one scoring function for this scenario is defined as:
\begin{align*}
  \lambda_{s_{in}} =\begin{cases}
    1, & \text{if $l_2 \neq l_1$}\\
    0, & \text{$Otherwise$} \numberthis
  \end{cases}.
  \label{eq2}
\end{align*}

We consider the synonyms of $\mathcal{M}(y)$ to further diversify the scoring function. Specifically, we use Wordnet \cite{miller1995wordnet} to obtain synonyms for $\mathcal{M}(y)$. We replace $\mathcal{M}(y)$ by its synonym to obtain a new sentence $s''_{in}$. Let the label predicted for $s''_{in}$ be $l_3$. The scoring function for this scenario is defined as:
\begin{align*}
  \lambda_{s_{in}} =\begin{cases}
    1, & \text{if $l_3 = l_1$}\\
    0, & \text{$Otherwise$}
  \end{cases}. \numberthis
  \label{eq4}
\end{align*}
Similarly, we can also consider the synonyms of $\mathcal{M}(y')$. 
The predicted label should flip if $\mathcal{M}(y)$ is replaced by synonyms of $\mathcal{M}(y')$.

Let $Z$ be the set of new sentences obtained through synonym replacement. The overall score for a given prompt ($P$) is defined as:
\begin{align*}
    Score(P) = \sum_{i=1}^{|S_{W}|} \sum_{j=1}^{|Z|} \lambda_{s_{ij}}. \numberthis
    \label{eq5}
\end{align*}
A higher score indicates that the prompt is more sensitive to the polarity of mapping tokens.

The score is calculated for all the prompts generated in the prompt augmentation step (Section \ref{prompt_augmentation}), and then the prompts are ranked based on their calculated score. The top-ranked prompt is the prompt with the highest score. Figure \ref{fig:ranking_metric} depicts the functioning of our ranking metric.

\subsection{Prediction}
\label{inference}

First, we define how we obtain the prediction probabilities using any given prompt. Given an input mapping $\mathcal{M}:\mathcal{Y}\rightarrow\mathcal{V}$ that maps the task label space to individual words in the vocabulary $\mathcal{V}$ of pre-trained MLM model $\mathcal{L}$, the probability of a label $y \in \mathcal{Y}$ for a given sentence $s_{in}$ in the unlabeled corpus $\mathcal{D}$ using a prompt $P$ is obtained as:
\begin{align*}
    p(y|s_{in}) = p([MASK] = \mathcal{M}(y) | s_{P}) \\ = \frac{exp(w_{\mathcal{M}(y)}.h_{[MASK]})}{\sum_{y' \in \mathcal{Y}} exp(w_{\mathcal{M}(y')}.h_{[MASK]})}, \numberthis
    \label{eq1}
\end{align*}
where $s_{P} = P(s_{in})$ is the sentence $s_{in}$ joined with the prompt $P$, which contains exactly one masked token at the position of the label, $h_{[MASK]}$ is the hidden vector of the [MASK] token and $w_v$ is the pre-softmax vector corresponding to $v \in \mathcal{V}$. The predicted label for the given sentence $s_{in}$ is the label $y$ with the highest probability.

Our proposed approach is to use quality prompts for the zero-shot prediction tasks. We can either select the top-ranked prompt or aggregate top-k-ranked prompts. If the top-1 prompt is selected, Eq. (\ref{eq1}) is used to obtain the label probability for each sentence, and the label with the highest probability is the predicted label.

Prompt aggregation may help correct the mistakes of the individual prompts. We consider prediction confidence and use the soft labels computed by Eq. (\ref{eq1}) in aggregation. Let $p_1(y), p_2(y), .., p_k(y)$ be the prediction probability for label $y \in \mathcal{Y}$ obtained using top-k prompts. The aggregated prediction probability is:
\begin{align*}
    p(y) = \frac{\sum_{i=1}^{k} Score(p_i)*p_i(y)}{\sum_{i=1}^{k} Score(p_i)}, \numberthis
    \label{eq7}
\end{align*}
and then the label with the highest aggregated prediction probability is chosen for the sentence.

\section{Experiments}

In this section, we evaluate the proposed \our model on several benchmark binary sentence-level sentiment classification datasets from various domains. More studies can be found in the Appendix \ref{sec:appendix}.

\subsection{Dataset}
\label{dataset}

\begin{table}[t]
\caption{ Statistics of the Datasets} 
\resizebox{\columnwidth}{!}{%
\begin{tabular}{l|l|l|l|l|l|l}
\hline
\multirow{2}{*}{\textbf{Datasets}} & \multicolumn{2}{c|}{\textbf{SST-2}} & \multicolumn{2}{c|}{\textbf{MR}}         & \multicolumn{2}{c}{\textbf{CR}} \\ \cline{2-7}
  & \textbf{Pos} & \textbf{Neg} & \textbf{Pos} & \textbf{Neg} & \textbf{Pos} & \textbf{Neg} \\ \hline

Train & 3610  & 3310  & 4331  & 4331  & 1407  & 368  \\ \hline
Dev & 444  & 428  & 0  & 0  & 0  & 0 \\ \hline
Test & 909 & 912 & 1000 & 1000 & 1000  & 1000 \\ \hline
\textbf{Total} & 4963 & 4650 & 5331 & 5331 & 2407  & 1368 \\ \hline
\end{tabular}
}
\label{table: Datasets Statistics}
\end{table}

The performance of \our is evaluated across 3 widely used sentiment classification datasets: SST-2~\cite{socher2013recursive}, MR~\cite{pang2002thumbs}, and CR~\cite{hu2004mining}. The dataset statistics are provided in Table \ref{table: Datasets Statistics}.

\subsection{Evaluation Metrics}
Since no training data is used in zero-shot settings, we evaluate all prompts on the \textit{entire dataset}. We use \textbf{Accuracy (Acc.)} and \textbf{macro F1 score (F1)} for all the datasets to evaluate the performance of \our and compare it with baselines under different settings. Note that Accuracy is equivalent to micro F1 score in binary classification tasks.

\begin{table*}[t]
\caption{Results of the sentiment classification task on the three benchmark datasets using BERT base and BERT large. We report accuracy and F1 score for all datasets. The results are evaluated on the entire dataset. We report the majority voting results for the automatic prompt baselines. The best-performing and runner-up model per column are
highlighted in bold and underlined, respectively.}
\centering
\resizebox{2\columnwidth}{!}{%
\begin{tabular}{c|c|c|c|c|c|c|c|c|c|c|c|c|c}
\hline
\multirow{3}{*}{\textbf{Method}} & \multirow{3}{*}{\textbf{Prompt}} &  \multicolumn{6}{c|}{\textbf{BERT base}} & \multicolumn{6}{c}{\textbf{BERT large}}\\ \cline{3-14}
& & \multicolumn{2}{c|}{\textbf{SST-2}} & \multicolumn{2}{c|}{\textbf{MR}} & \multicolumn{2}{c|}{\textbf{CR}} & \multicolumn{2}{c|}{\textbf{SST-2}} & \multicolumn{2}{c|}{\textbf{MR}} & \multicolumn{2}{c}{\textbf{CR}} \\
\cline{3-14}
 &   & \textbf{Acc.} & \textbf{F1} & \textbf{Acc.} & \textbf{F1} & \textbf{Acc.} & \textbf{F1} & \textbf{Acc.} & \textbf{F1} & \textbf{Acc.} & \textbf{F1} & \textbf{Acc.} & \textbf{F1}\\
\hline
LM-BFF & \multirow{2}{*}{Automatic} & 58.46 & 62.24 & 57.94 & 62.81 & 71.35 & 69.66 & 52.69 & 59.33 & 57.3 & 63.69 & 70.55 & 69.11 \\
UPT & & 57.46 & 61.79 & 62.65 & 66.78 & 75.09 & 73.53 & 53.82 & 61.08 & 65.2 & 69.69 & 72.62 & 71.4 \\ \hline 
LM-BFF & \multirow{4}{*}{Manual} & 62.3 & 65.75 & 58.18 & 62.16 & 74.9 & 72.81 & 61.15 & 65.41 & 57.88 & 62.64 & 72.59 & 70.85 \\
PPT & & 52.53 & 56.93 & 50.5 & 53.41 & 64.03 & 61.02 & 52.29 & 57.68 & 50.5 & 56.0 & 63.9 & 62.21 \\
Base Prompt$\dagger$ & & 62.3 & 65.75 & 58.18 & 62.16 & 74.9 & 72.81 & 61.15 & 65.41 & 57.88 & 62.64 & 72.59 & 70.85 \\
Base Prompt$\star$ & & 63.22 & 63.15 & 59.97 & 60.25 & 69.04 & 64.29 & 54.12 & 58.6 & 54.43 & 57.12 & 56.59 & 62.14\\ \cline{1-1}\cline{2-14}
\textbf{\our (Top-1)$\dagger$} & \multirow{6}{*}{Automatic} & 67.48             & 67.52          & 58.93           & 62.07         & 73.36           & 70.16         & 74.13             & 75.66          & 69.84           & 71.75         & 73.12           & 70.65 \\
\textbf{\our (Top-3)$\dagger$} &  & 67.12             & 68.22          & 60.15           & 60.14         & 71.19           & 68.23         & 67.58             & 70.65          & 64.15           & 67.91         & 70.05           & 67.82 \\
\textbf{\our (Top-5)$\dagger$} & & 67.99             & 68.94          & 61.19           & 62.92         & 71.51           & 69.32         & 66.55             & 70.09          & 63.47           & 67.76         & 69.41           & 67.32 \\ \cline{1-1} \cline{3-14}
\textbf{\our (Top-1)$\star$} & & \textbf{72.18}             & \textbf{72.36}          & \textbf{68.24}         & \textbf{68.26}         & 75.09           & 72.1          & 74.74             & 74.71          & 70.29           & 70.36         & \underline{80.47}           & \underline{78.43} \\
\textbf{\our (Top-3)$\star$} & & \underline{71.92}             & \underline{72.01}          & \underline{67.88}           & \underline{67.89}         & \underline{76.82}           & \underline{74.43}         & \textbf{77.11}             & \textbf{77.58}          & \textbf{72.96}           & \textbf{73.54}         & 79.17           & 77.84  \\
\textbf{\our (Top-5)$\star$} & & 71.5              & 71.46          & 66.74           & 66.88         & \textbf{77.26}           & \textbf{74.52}        & \underline{76.9}              & \underline{77.54}          & \underline{72.46}           & \underline{73.43}         & \textbf{81.45}           & \textbf{79.52}\\
\hline
\end{tabular}
}
\label{table: bert_base_results}
\end{table*}
\subsection{Baseline Methods}
  
Since none of the prior work has performed the task of binary sentence-level sentiment classification under the zero-shot setting, we compare it with the baselines that have performed the task under the few-shot setting for the datasets discussed in Section \ref{dataset}. For a fair comparison, we modified these studies as per the zero-shot setting, using the prompts reported in their paper. The baseline templates are discussed in Table \ref{table: top_baseline} of Appendix \ref{sec:appendix}.

\textbf{LM-BFF} \cite{gao2021making}: This paper explores manual prompts and generates automatic prompts under the few-shot setting. Specifically, they use few-shot examples to automatically generate prompts using the T5 model. The performance of their method is evaluated on a range of classification and regression tasks using RoBERTa-large \cite{liu2019roberta} with fine-tuning. We compare \our with their manual prompt and their top-ranked automatic prompts.

\textbf{PPT} \cite{gu2022ppt}: This paper proposes pre-training hard prompts by adding soft prompts to achieve better initialization into the pre-training stage on classification tasks. \our is compared with their manual prompt.

\textbf{UPT} \cite{wang2022towards}: This paper proposes a Unified Prompt Tuning framework and designs prompts by fine-tuning a pre-trained language model (RoBERTa-large) over a series of non-target NLP tasks. After multi-task training, the trained model can be fine-tuned to fit unseen tasks. \our is compared with their top-ranked prompts.

\subsection{Settings}
The experiments are conducted using pre-trained uncased BERT (BERT base and BERT large) encoders. BERT base has $12$ attention heads, $12$ hidden layers, and a hidden size of $768$ resulting in $110$M pre-trained parameters, whereas BERT large has $16$ attention heads, $24$ hidden layers, and a hidden size of $1024$ resulting in $336$M pre-trained parameters. We set $K$, the hyperparameter for the number of candidate words in paraphrasing, to $30$. We obtain $6$ synonyms for each mapping word from WordNet \cite{miller1995wordnet}. The size of the set of new sentences through synonym replacement ($Z$) is $12$, $6$ of which are obtained by replacing the mapping token $\mathcal{M}(y)$ with its synonyms, and the other $6$ are obtained by replacing the mapping token by $\mathcal{M}(y')$ and synonyms of $\mathcal{M}(y')$, where $y \neq y'$. 

For \our, we considered two different base prompts. The first base prompt is \textit{"<sentence>. It was [MASK]"}, which is the same as the manual prompt used by LM-BFF (denoted by $\dagger$ in Table \ref{table: bert_base_results}), whereas the second base prompt is \textit{"<sentence>. The sentence was [MASK]"} (denoted by $\star$ in Table \ref{table: bert_base_results}). The base prompts defined are generic and used for all datasets.

\subsection{Results and Discussion}

To better compare the performance of different methods, we categorize them based on the prompt (manual or automatic).

Table \ref{table: bert_base_results} shows the results of all prompts using BERT base and BERT large pre-trained MLM models, respectively. \our with the $\star$ base prompt significantly outperforms both manual and automatic baseline methods on both pre-trained MLM models on all three datasets. Overall, the aggregation strategy tends to outperform the selection strategy, but the outperformance is inconsistent across different data. We conduct more studies on the impact of top-k prompts in Section \ref{sec:slc-vs-agg}.

\begin{table*}[t]
\caption{Ablation study results with and without WordNet on the three benchmark datasets for sentiment classification tasks. We report accuracy and F1 score for all datasets using BERT base and BERT large. The results are evaluated on the entire dataset.}
\centering
\small
\begin{tabular}{c|c|c|c|c|c|c|c}
\hline
\multirow{2}{*}{\textbf{Method}} & \multirow{2}{*}{\textbf{Encoder}} & \multicolumn{2}{c|}{\textbf{SST-2}} & \multicolumn{2}{c|}{\textbf{MR}} & \multicolumn{2}{c}{\textbf{CR}} \\
\cline{3-8}
 &  &  \textbf{Acc.} & \textbf{F1} & \textbf{Acc.} & \textbf{F1} & \textbf{Acc.} & \textbf{F1} \\
\hline
\our-W (Top-1) & \multirow{2}{*}{} & 62.77 & 64.14 & 59.25 & 63.3 & 72.04 & 71.29 \\
\our-W (Top-3) & & 62.57 & 65.73 & 60.1 & 64.34 & 75.78 & 72.76 \\ 
\our-W (Top-5) & BERT& 62.85 & 66.41 & 61.0 & 64.91 & 75.67 & 73.63 \\ \cline{1-1}\cline{3-8}
\our (Top-1) & base& 72.18 & 72.36 & 68.24 & 68.26 & 75.09 & 72.1 \\
\our (Top-3) & \multirow{2}{*}{}& 71.92 & 72.01 & 67.88 & 67.89 & 76.82 & 74.43 \\ 
\our (Top-5) & & 71.5 & 71.46 & 66.74 & 66.88 & 77.26 & 74.52 \\
\hline \hline
\our-W (Top-1) & \multirow{2}{*}{} & 73.55 & 74.1 & 70.29 & 70.36 & 80.47 & 78.43 \\
\our-W (Top-3) & & 74.54 & 75.0 & 69.94 & 71.03 & 79.17 & 77.83 \\ 
\our-W (Top-5) & BERT& 75.68 & 76.74 & 71.89 & 73.14 & 81.0 & 78.94 \\ \cline{1-1}\cline{3-8}
\our (Top-1) & large& 74.74 & 74.71 & 70.29 & 70.36 & 80.47 & 78.43 \\
\our (Top-3) & \multirow{2}{*}{}& 77.11 & 77.58 & 72.96 & 73.54 & 79.17 & 77.84 \\ 
\our (Top-5) & & 76.9 & 77.54 & 72.46 & 73.43 & 81.45 & 79.52 \\
\hline
\end{tabular}
\label{table: bert_ablation_study}
\end{table*}

It is interesting to notice that for $\dagger$ base prompt \our outperforms on SST-2 and MR datasets but not on the CR dataset. Furthermore, the margin of \our over the base prompt decreases for $\dagger$ compared to $\star$ base prompt. This is because ``It was'' is harder to augment than ``The sentence was'' since the former is shorter and contains no concrete word. Even though the $\dagger$ base prompt is not ranked top-1 by \our on the CR dataset, it is ranked as the $4$-th for both pre-trained MLM models, demonstrating that \our can recognize $\dagger$ base prompt as a high-quality prompt.

It is also interesting to note that for baseline methods, either using manual or automatic prompts, there is no significant gain using the BERT large over the BERT base encoder, and the performance of a prompt can change significantly using different pre-trained language models. However, we can observe that the performance of \our improves with the scale of the model. The key difference between \our and the automatic prompts generated by baseline models is that we use the same language models to generate prompts and conduct classification tasks, whereas baselines generate prompts manually or using a different model. These results suggest that different language models have different knowledge of the language, so prompts need to be generated specifically for the chosen language model. 

\subsection{Study of Selection VS Aggregation}\label{sec:slc-vs-agg}

Comparing top-1 selection to top-k aggregation, from Table \ref{table: bert_base_results}, we can observe that top-1 selection performs better compared to top-k aggregation on BERT base whereas on BERT large top-k aggregation performs better. Furthermore, we can observe that the top-k aggregation result does not increase with k as suggested by previous works \cite{gao2021making}.

\begin{figure}[t]
    \centering
    \includegraphics[width=0.42\textwidth]{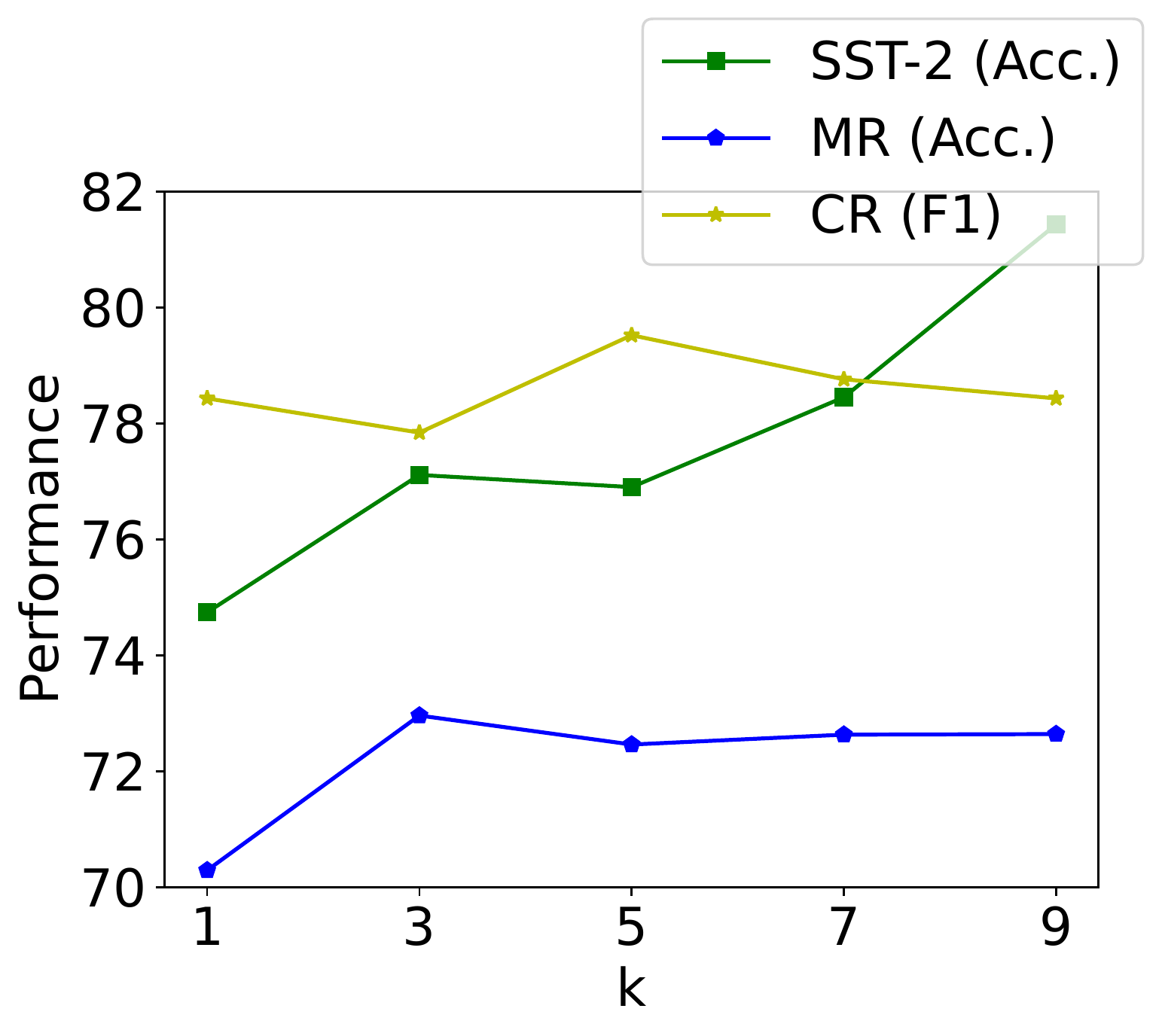}
    \caption{Performance vs the number of aggregated top-k prompts for BERT large on $\star$ base prompt.}
    \label{fig:aggregation_plot}
\end{figure}

To further analyze our observation, we plot the change in performance of \our with respect to the number of aggregated top-k prompts for BERT large encoder on $\star$ base prompt in Figure \ref{fig:aggregation_plot}. Figure \ref{fig:aggregation_plot} shows that the top-k aggregation performance increases with $k$ only for SST-2 dataset and does not increase for CR and MR datasets. This implies that top-k aggregation performance increases with $k$ only for some datasets but not all. Furthermore, we can also observe that top-k aggregation performance can be better than top-1 selection performance on all three datasets. We believe that aggregation performance improves when the top-ranked prompts make independent mistakes. 

\subsection{Study of the Proposed Ranking Metric}
\begin{figure}[t]
    \centering
    \includegraphics[width=0.40\textwidth]{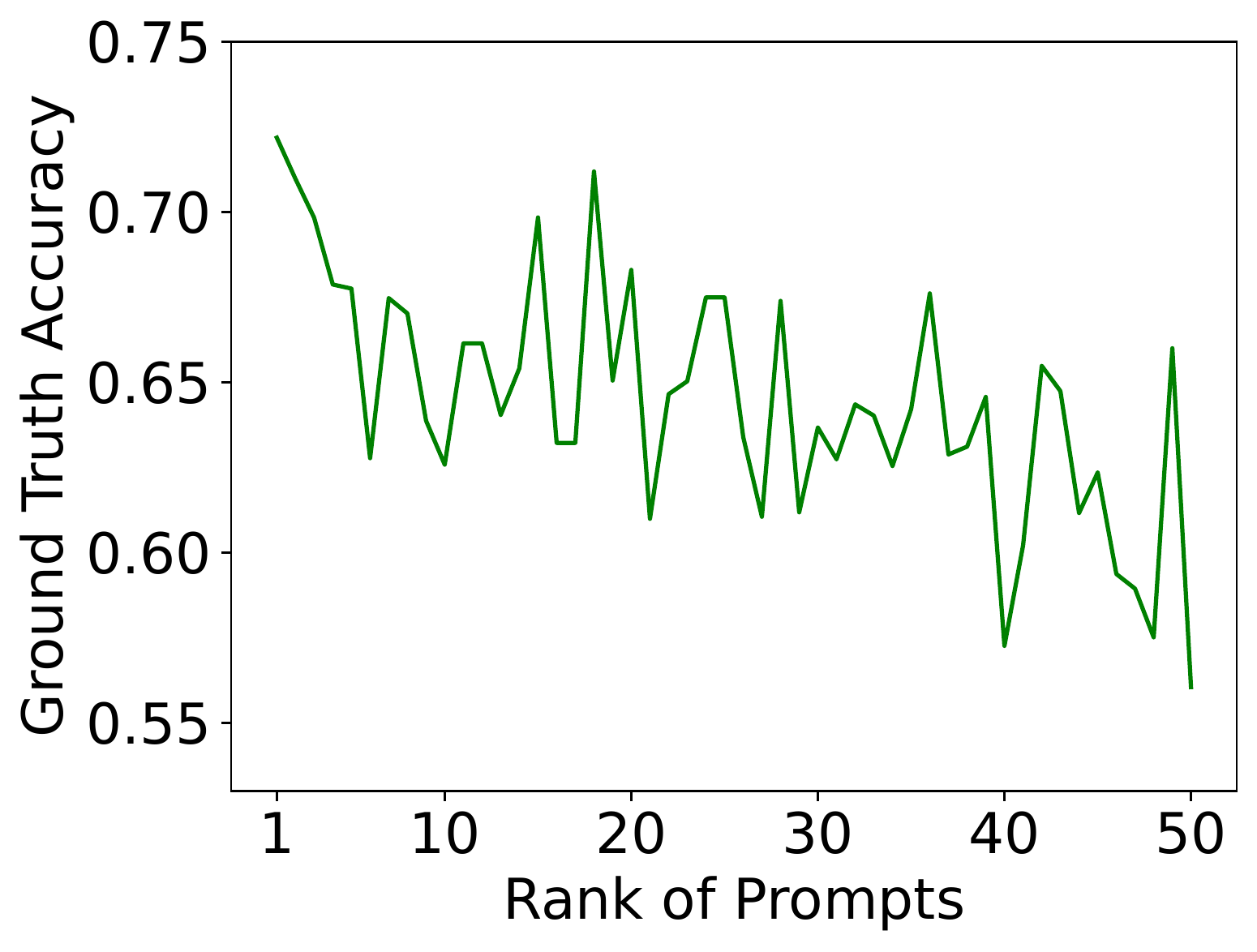}
    \caption{Proposed Metric vs Ground Truth Performance for SST-2 dataset using BERT base model on $\star$ base prompt .}
    \label{fig:metricvsgt}
\end{figure}

To study the effectiveness of the proposed ranking metric, we plot the accuracy of the augmented prompts evaluated using ground truth labels with respect to their ranks based on the proposed ranking metric. The results for SST-2 dataset using the BERT base model on $\star$ base prompt are shown in Figure \ref{fig:metricvsgt}. The figure shows that the highly-ranked prompts achieve higher accuracy than the low-ranked prompts in general, demonstrating the effectiveness of our proposed ranking metric. Furthermore, we can observe that the accuracy of the prompts decreases as the rank provided by our proposed ranking metric increases. 

\subsection{Ablation Studies}

We conduct ablation studies to investigate the contributions of Wordnet synonyms to the overall model performances.

Table \ref{table: bert_ablation_study} shows the performance of \our with and without Wordnet. From the results, we can observe that \our with Wordnet outperforms \our without Wordnet for both variants of pre-trained MLM models. The results show that diversification of the mapping tokens helps the scoring function to rank the prompts better and subsequently improve the performance.

\section{Conclusion}

This work proposes to study how to improve manual prompts for binary sentence-level sentiment classification tasks under zero-shot settings. To overcome the sensitivity of the language model to a manual prompt, we propose prompt augmentation techniques to generate multiple candidate prompts. Further, to rank the generated prompts without labeled data, we propose a novel ranking metric based on the intuition that high-quality prompts should be sensitive to the change of certain keywords in the given sentence. Extensive experiments and ablation studies demonstrate the power of the proposed \our on three benchmark datasets.

\section*{Limitations}
The proposed method is tested for a binary labeling scenario where each instance can belong to one of the labels but not both. The scenario of overlapping labeling space is not tested, nor is the scenario for multi-class labeling space. Since we aim to obtain high-quality prompts similar to the base prompt, if the base prompt is very restrictive, then the suggested prompt might be the same as the base prompt. The approach only applies to two moderately sized MLM models, and the extension to other larger models is not tested.
\section*{Ethics Statement}

We comply with the ACL Code of Ethics. 

\bibliography{anthology,custom}
\bibliographystyle{acl_natbib}
\clearpage
\appendix

\section{Appendix}
\label{sec:appendix}

\subsection{Case Study}
\label{case_study}

\begin{table*}[t]
\caption{Top 5 Ranked Prompts for BERT large and BERT base}
\centering
\small
\begin{tabular}{c|p{7cm}|p{6cm}}
\hline
\textbf{Dataset} & \textbf{BERT large} & \textbf{BERT base} \\
\hline
\multirow{5}{*}{\textbf{SST-2}} & The sentence sounded [MASK] because <sentence> . & <sentence>. Every sentence was [MASK] . \\
& Every sentence was [MASK] . <sentence> . & Every sentence was [MASK]. <sentence> . \\
& <sentence> . Every sentence was [MASK] . & Each sentence was [MASK] . <sentence> . \\
 & The result was [MASK] . <sentence> . & <sentence>. Each sentence was [MASK] . \\
   & Each sentence was [MASK] . <sentence> . & <sentence> so every sentence was [MASK] . \\
     \hline
\multirow{5}{*}{\textbf{MR}} & The sentence sounded [MASK] because <sentence> . & <sentence>. Every sentence was [MASK] . \\
 & The sentence seemed [MASK] because <sentence> . & Every sentence was [MASK]. <sentence> .\\
  & The result was positive . <sentence> . & Each sentence was [MASK] . <sentence> .\\
   & Every sentence was [MASK] because <sentence> . & <sentence> . Each sentence was [MASK] . \\
    & Every sentence was [MASK] . <sentence> . & <sentence> so the sentence sounded [MASK] . \\ \hline
\multirow{5}{*}{\textbf{CR}} & The sentence sounded [MASK] because <sentence> . & The sentence sounded [MASK] . <sentence> . \\
 & The sentence sounded [MASK] . <sentence> . & <sentence> . The sentence sounded [MASK] . \\
  & <sentence> . The sentence sounded [MASK] . & Every sentence was [MASK] . <sentence> . \\
  & Every sentence was [MASK] . <sentence> . & <sentence> . Every sentence was [MASK] . \\
  & The answer was [MASK] . <sentence> . & This sentence was [MASK] . <sentence> . \\ \hline
\end{tabular}
\label{table: top_5_ranked}
\end{table*}

\begin{table*}[t]
\caption{Ranked Prompts of Baselines}
\centering
\small
\begin{tabular}{c|p{5cm}|p{3cm}|p{5cm}}
\hline
\textbf{Dataset} & \textbf{LM-BFF} & \textbf{PPT} & \textbf{UPT} \\
\hline
\multirow{5}{*}{\textbf{SST-2}} & <sentence>. A [MASK] one. & <sentence>. [MASK]. & <sentence>. It was [MASK]. \\
& <sentence>. A [MASK] piece. &  & <sentence>. I thought it was [MASK]. \\
& <sentence>. All in all [MASK]. & & <sentence>. It is [MASK]. \\
&  & & <sentence>. The review is [MASK]. \\
&  & & <sentence>. A [MASK] one.\\
     \hline
\multirow{4}{*}{\textbf{MR}} & It was [MASK] ! <sentence>.  & <sentence>. [MASK]. & <sentence>. A [MASK] piece of work. \\
& <sentence>. It’s [MASK]. & & <sentence>. It is [MASK]. \\
& <sentence> A [MASK] piece of work. & & <sentence>. The film is [MASK].\\
&  & & <sentence>. A really [MASK] movie. \\ 
     \hline
\multirow{5}{*}{\textbf{CR}} & <sentence>. It’s [MASK] ! & <sentence>. [MASK]. & <sentence>. It was [MASK].\\
 & <sentence>. The quality is [MASK]. & & <sentence>. It looks [MASK]. \\
& <sentence>. That is [MASK]. & & <sentence>. It is [MASK]. \\
& & & <sentence>. The quality is [MASK].\\
& & & <sentence>. I thought it was [MASK]. \\ \hline
\end{tabular}
\label{table: top_baseline}
\end{table*}

Table \ref{table: top_5_ranked} shows the top-5 ranked prompts for three datasets, SST-2, MR, and CR. The table shows that prompts with subordinate conjunctions like \textit{``because"} and \textit{``so"} are ranked higher. The ranking confirms our intuition that subordinate conjunctions that introduce a dependency between the prompt and the sentence can improve the performance of the prompts. Note that the proposed ranking metric ensures that low-quality prompts are not ranked higher. Therefore the results from the table suggest that prompts with subordinate conjunctions are high-quality.

\end{document}